\documentclass[aps,reprint,groupedaddress,amsmath,amssymb]{revtex4-1}

\usepackage{amsmath,amssymb,bm,amsthm}
\usepackage{graphicx,xcolor}
\usepackage[format=plain,justification=justified,singlelinecheck=false]{caption}
\usepackage[position=top]{subfig}
\usepackage[bookmarks=false]{hyperref} 
\usepackage[nameinlink,capitalise]{cleveref}
\crefname{equation}{}{}
\let\ref\cref
\let\eqref\cref
\let\autoref\cref
\hypersetup{
  colorlinks   = true, 
  urlcolor     = blue, 
  linkcolor    = blue, 
  citecolor   = magenta 
}

\newcommand{\p}{\partial}
\newcommand{\rar}{\rightarrow}
\newcommand{\J}{\mathcal{J}}
\newcommand{\D}{\mathcal{D}}
\newcommand{\M}{\mathcal{M}}
\usepackage{mathtools}
\newcommand\myeq{\stackrel{\scriptsize\mathclap{\mbox{def.}}}{=}}

\graphicspath{{./}{./figs/}{./figs/nets/}{./figs/gap-tooth/}}

\begin{document}
 

\title{Particles to Partial Differential Equations Parsimoniously}


\author{Hassan Arbabi}
\email[]{arbabiha@gmail.com}
\affiliation{ Department of Mechanical Engineering, MIT
}

\newif\ifcomments
\newcommand\comments[1]{\ifcomments\textcolor{cyan}{ {HA:  \textbf{#1}}}\else\relax\fi}
\commentstrue

\newif\ifchanged
\newcommand\changed[1]{\ifcomments\textcolor{red}{ #1}\else\relax\fi}
\commentstrue

\author{Ioannis G. Kevrekidis }
 \email{yannisk@jhu.edu}
\affiliation{%
 Department of Chemical and Biomolecular Engineering, Johns Hopkins University}

\date{\today}

\begin{abstract}
Equations governing physico-chemical processes are usually known at microscopic spatial scales, yet one suspects that there exist equations, e.g. in the form of Partial Differential Equations (PDEs), that can explain the system evolution at much coarser, meso- or macroscopic length scales. 
Discovering those \emph{coarse-grained effective} PDEs can lead to considerable savings in computation-intensive tasks like prediction or control.
We propose a framework combining artificial neural networks with multiscale computation, in the form of equation-free numerics,  for efficient discovery of such macro-scale PDEs directly from microscopic simulations. Gathering sufficient microscopic data for training neural networks can be computationally prohibitive;  equation-free numerics enable a more parsimonious collection of training data by only operating in a sparse subset of the space-time domain. 
We also propose using a data-driven approach, based on manifold learning and unnormalized optimal transport of distributions, to identify macro-scale dependent variable(s) suitable for the data-driven discovery of said PDEs. This approach can corroborate physically motivated candidate variables, or introduce new data-driven variables,  in terms of which the coarse-grained effective PDE can be formulated. 
We illustrate our approach by extracting coarse-grained evolution equations from particle-based simulations with \emph{a priori} unknown macro-scale variable(s), while significantly reducing the requisite data collection computational effort.
\end{abstract}

\maketitle

\section{Data-driven discovery of coarse-grained PDE{s} \label{sec_intro}}
In many physical, chemical or biological systems, a description of the system temporal evolution is available at a fine scale: atoms on a crystal lattice, response of individual neurons in a tissue to stimuli, 
individual cell dynamics in chemotaxis, reactive molecular dynamics. Yet, using this type of model for
computation and analysis at a macroscopic, engineering level will be prohibitive, since it requires processing interactions between enormous numbers of fine-scale components. 
The traditional approach to resolving this problem is to obtain models of the system at the macroscopic level, that is, (a) identify collective variables that describe statistical properties of the fine-scale components (e.g. concentration of species rather than position of individual atoms), and (b) obtain  governing equations for these macro-scale variables (empirically or mathematically) from the fine-scale description.  
The closures required to obtain these equations typically require extensive expertise,  several assumptions (for closed-form mathematical derivations) and/or large bodies of experiments.
This limits the class of systems for which coarse-graining is readily applicable. 

An emerging paradigm for discovery of macro-scale governing equations is based on data-driven learning.
The recent explosive progress in Machine Learning has empowered many algorithmic approaches to the discovery of model PDEs either in a fully data-driven fashion  \cite[e.g.][]{ krischer1993model,gonzalez1998identification,rudy2017data,vlachas2018data,pathak2018model,lu2019deeponet,linot2020deep} or in a physics-informed data-assisted manner, \cite[e.g.][]{raissi2018hidden,wang2017physics,raissi2019physics,psichogios1992hybrid,rico1994continuous,bar2019learning}.  In particular, the recent work in \cite{lee2020coarse} has utilized ML algorithms 
to identify relevant macro-scale variables as well as approximate the associated coarse-scaled PDEs from fine-scale data. Application of this framework to a (fine-scale) lattice Boltzmann model produced a ``hydrodynamic-level" PDE model comparable, in prediction performance, to established closed-form PDE models of the coarse-grained system. 

Despite the promise of ML methods, two persistent obstacles limit the discovery of coarse-grained PDEs from fine-scale data:
The first is the need to generate sufficient amounts of fine-scale data, required to train the ML coarse model; the quantity of necessary data grows with the complexity of the dynamics. The second obstacle is identifying  the right macro-scale variable(s) for which an appropriate PDE can be learned. Here we present a framework mitigating these two obstacles.

We link machine learning of coarse-scale PDEs with equation-free multiscale computation \cite{kevrekidis2003equation,gear2003gap} which leads to parsimonious generation of required training data. This is accomplished by utilizing the hypothesized existence of macro-scale dynamics  to \emph{simulate the microscopic model in only a small fraction of the space-time domain}. We also present a framework for data-driven identification of macro-scale variables  for microscopic systems described by agent-based or particle-based models. This framework is based on data-mining the fine-scale simulation results. In particular, we apply Diffusion Maps \cite{coifman2006diffusion}  to observations of the fine scale simulation (local snapshots of particle distributions) and use the notion of \emph{unnormalized optimal transport distances} between those distributions \cite{chizat2018scaling,gangbo2019unnormalized}. 

We then use simple neural network architectures to discover the coarse-grained effective PDE laws from the fine-scale data using either physically motivated variables (if known) or the ones discovered from data when necessary. We look for two representations of the PDEs: in the first formulation, we feed the neural net the value of the macro-scale variable and its spatial derivatives on a macro-space grid, and learn the time derivative(s) at each grid point. In the second formulation, we directly learn the discretized PDE on a spatial grid; that is, we feed the neural net with the local values of the macro-scale variable \emph{ on a local stencil}, and train the network to approximate the time derivative on the grid points. We show the promise of our framework by learning a viscous Burgers PDE from (parsimonious) particle simulation data. 

The paper is organized as follows: we first describe the problem setup using a particle-based model whose coarse-grained PDE is known. Then we discuss the data mining approach for discovering the PDE variable from data. Next, we describe the PDE learning approach and the architecture  of the neural networks used for learning. We present learning results using both the known macro-scale variable and the data-driven variable. We close with a brief discussion of the limitations of the approach and promising future directions. 
Source codes and data outlining our application of this framework are available at  \url{https://github.com/arbabiha/Particles2PDEs}.

\section{Setup: a particle-based model \label{sec_setup}}
Consider a system with available microscopic-level description: an atomistic, agent-based or lattice model. We assume there exists a macro-scale spatial field description for the system, in terms of one or more collective variables, evolving in time according to an unknown PDE.  Throughout this paper, we use the following example of a particle-based model for concrete presentation of ideas:
our one-dimensional microscopic model of particle motion is given by
\begin{equation}\label{eq_Burgersmicro}
X_p(n+1)=X_p(n) + \frac{mh}{Z d_{p,m}} + \sqrt{2\nu h} W_n
\end{equation}
where $X_p$ is the location of the $p$-th particle, $n$ is the time index,  $\nu$ and $h$ are parameters of the motion, and $W_n$'s are independent random variables with standard normal distribution. The term $m/Z d_{p,m}$ denotes the coupling between particles: $d_{p,m}$ is the distance between $m$-th particle to the right and to the left of our $p$-th particle, and the parameter $Z$ regulates the strength of this coupling. 
We assume particles move on the periodic domain $[0,2\pi)$.

Let $\rho(x,t)$ denote the particle density field in this particle model. 
As discussed in \cite{li2007deciding}, the model assigns each particle with a drift speed of (a local estimate of) $\rho/2$, leading to a coarse-grained flux of $\rho^2/2-\nu \p_x\rho$. Coarse graining of the particle model in \cref{eq_Burgersmicro} will thus lead to the well-known Burgers equation:
\begin{equation}\label{eq_BurgersPDE}
 \p_t\rho = - \rho\p_x\rho + \nu \p_{xx} \rho,
\end{equation}
where the parameter $\nu$ is called kinematic viscosity. 
In passing from the particle model to the Burgers PDE, the resolution factor is interpreted as the the number of particles in a unit mass, i.e.,
\begin{equation}\label{eq_resfactor}
 Z = \frac{\text{total no. of particles}}{\int_{0}^{2\pi} \rho dx}.
\end{equation}
  and $h$ as the length of time steps in the particle motion.
Comparison of our learned PDE results to the Burgers PDE will validate our framework.

In the above example, the macro-scale variable, for which the evolution PDE can be expressed,  is known (i.e. the particle density field), and the transformation from microscopic to macro-scale variable (the ``restriction") is straightforward: given a position in the domain, we compute the local particle histogram to get an approximation of the density field.

To simulate the micro-scale system we start from a prescribed macro-scale initial condition $\rho_0(x)$. First, we set a resolution factor, and \emph{lift} the initial macro-scale $\rho_0$ to a consistent micro-scale state: we partition the domain into $N$ equal-sized intervals ($\Delta x=2\pi/N$) and compute the number of particles within the $i$-th interval as
\begin{equation}
n_i =\rho_0(x_i) \Delta x Z
\end{equation}
where $x_i$ is the center of the $i$-th interval.  We insert $n_i$ particles, with random positions drawn from the uniform distribution, inside this interval. 
In a standard micro-scale simulation, we evolve the microscopic model in \cref{eq_Burgersmicro} for all the particles and record their final position after a time interval. Then we evaluate the macro-scale state, i.e., the density field, through local histograms of the new particle positions. 

\subsection{Parsimonious generation of training data via the gap-tooth scheme \label{sec_gaptooth}}
To learn the macro-scale PDE for the particle model we must generate a large number of snapshots. Interestingly, even when we do not know this macro-scale PDE explicitly, we can use its conceptual existence to substantially reduce the computational effort for the micro-scale simulations.  This idea is the basis of the equation-free multiscale approach \cite{kevrekidis2003equation}.   Here, we propose to use this approach, already established in the scientific computation literature, for parsimonious generation of fine-scale training data.
Our micro-scale simulations are performed inside a sparse grid of boxes that only cover a fraction of the space-time domain. The smoothness of the macro-scale variable is used to \emph{couple} the micro-scale simulations within neighboring boxes.  This coupling allows the global solution to emerge from the partial domain simulations. Here, we only briefly review (and then use) a particular equation-free algorithms, the so-called \emph{gap-tooth scheme} \cite{gear2003gap}, and refer the reader to  \cite{kevrekidis2003equation,samaey2005gap,roberts2004higher}  for details and analysis.

\begin{figure}
\subfloat[][]{\includegraphics[scale=1]{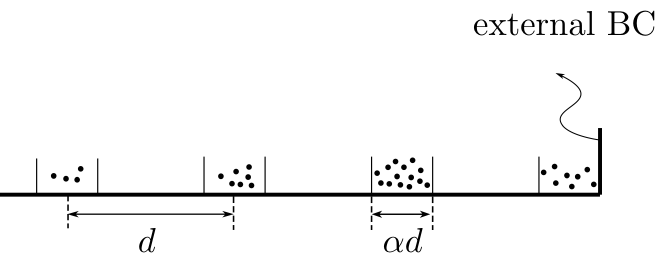}} 

\subfloat[][]{\includegraphics[scale=1]{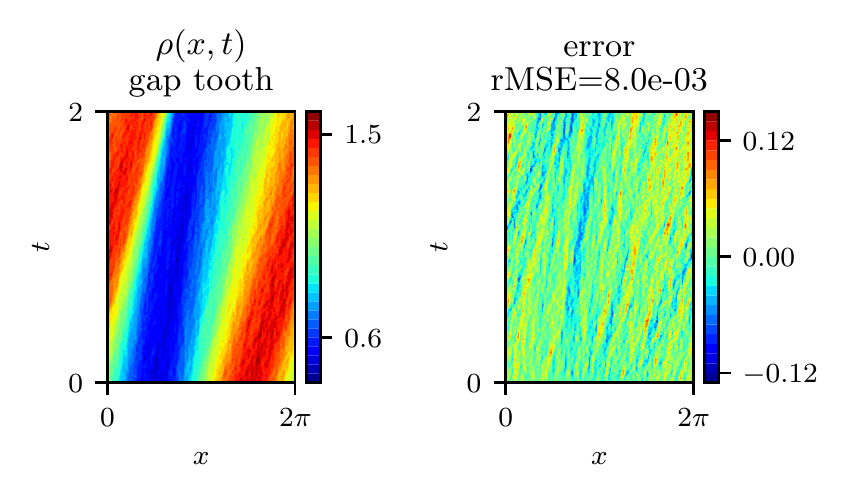}} 

\subfloat[][]{\includegraphics[scale=1]{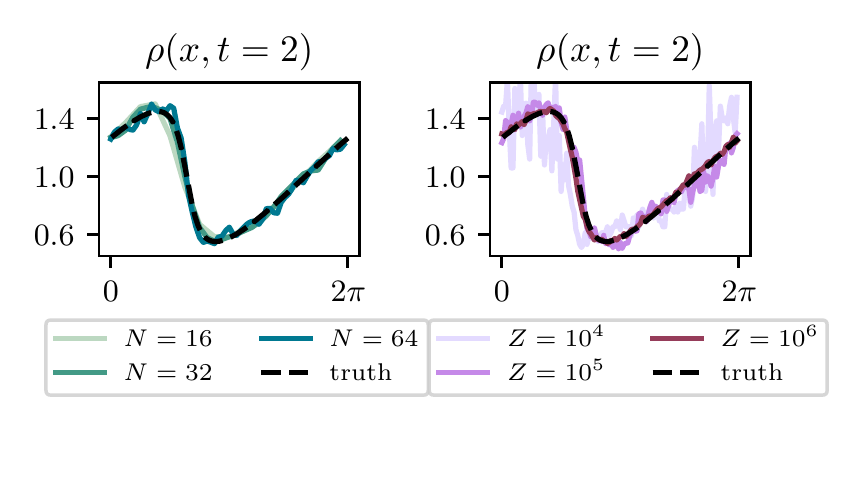}} 

\caption{\textbf{Gap-tooth scheme for simulation of the microscopic Burgers model.} a) The gap-tooth grid, consisting of teeth and gaps in-between. The particle simulations are carried out only inside the teeth. b) Gap-tooth simulation vs. ``truth" for the particle model in \cref{eq_Burgersmicro} ($\alpha=0.1, Z=10^4, N = 32$) c)  Macro-snapshot of the  micro-Burgers simulation solution at the final time: dependence on the number of teeth ($N$) with fixed $Z=10^5$ (left), and dependence on the resolution factor $Z$ with fixed $N=128$ (right). }
\label{fig_gaptooth}
\end{figure}

In the gap-tooth scheme the spatial domain is populated 
with a grid of boxes (the ``teeth")  covering a fraction ($\alpha$) of the spatial domain, as shown in \cref{fig_gaptooth}(a).  For simplicity, we assume that the grid is uniformly spaced  and uniformly sized, with $d$ and $\alpha d$ denoting the distance of two adjacent teeth and each tooth width, respectively. We lift the initial density profile $\rho_0$ to a particle state within each tooth using a process similar to the one described above. 
After setting up the grid and lifting of the initial density, we simulate the microscopic dynamics in \eqref{eq_Burgersmicro} for all particles over a short time period. If a particle jumps out of a tooth during this period, it is redirected to one of the neighboring teeth using flux redistribution laws derived from interpolations of macro-scale fluxes.
For more details see \cref{app_gaptooth} or \cite{gear2003gap}.

\Cref{fig_gaptooth}(b) shows an example of the gap-tooth multiscale simulation for the Burgers model with initial density profile $\rho(x,t=0)=1-0.5\sin x$ and parameter value $\nu=0.05$. Particle simulations are carried out in only 10\% of the space with $Z=10^4$. 
The solution at $t=2$ for various teeth grid sizes and resolution factors is shown in \cref{fig_gaptooth}(c). Increasing the resolution factor, the density estimates become more consistent; increasing the grid size leads to less bias.

\section{Discovering the macro-scale variable \label{sec_variableID}} 
The first question arising in discovering the macro-scale equations from microscopic data is the choice of macro-scale variable to be used in the modeling.
 Here, we propose using data mining on snapshots of fine-scale data, obtained from gap-tooth simulations, to  identify an appropriate macro-scale variable.
 In the case of our particle model, we show that our approach leads to a macro-scale (collective) variable which is one-to-one with the particle density field.

\begin{figure*}
\centering
\subfloat[][]{\includegraphics[scale=1]{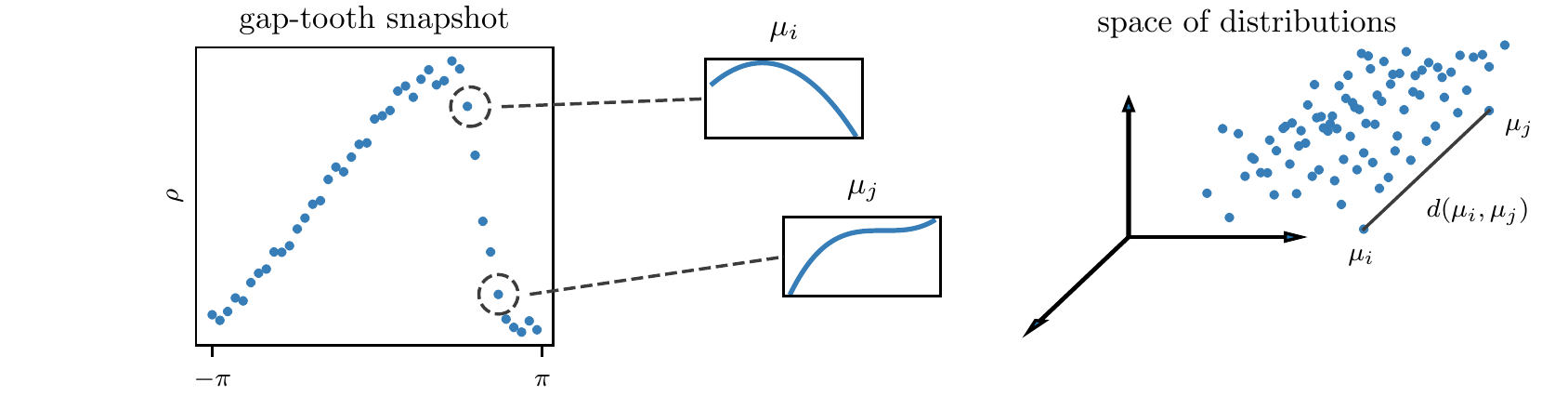}} 
 
\subfloat[][]{\includegraphics[scale=1]{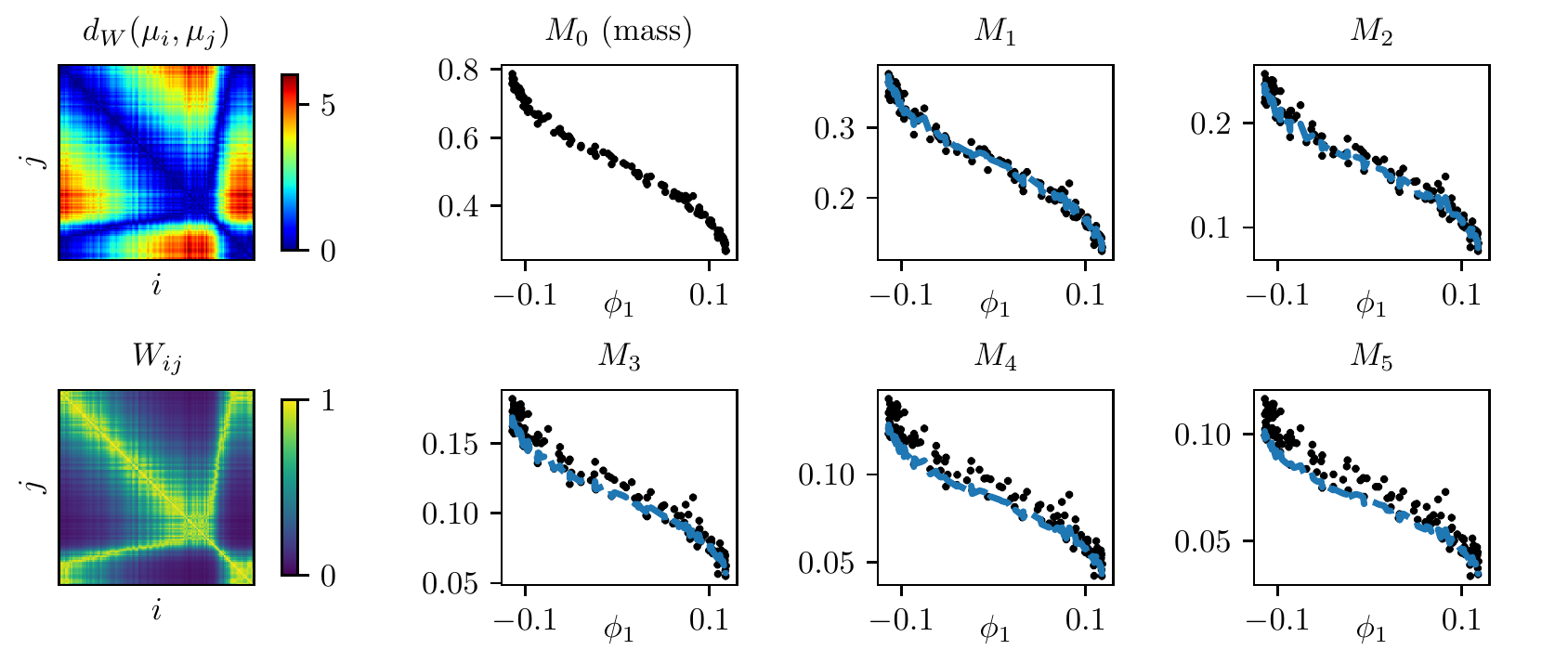}}


\caption{ \textbf{Identification and modeling of the macro-scale variable:} a) The particle distribution within each tooth is extracted from snapshots of gap-tooth simulation, b) The pairwise distances between these distributions are computed using unbalanced optimal transport ($\beta=1$), and fed to the Diffusion-Maps algorithm to estimate the macro-scale variable $\phi_1$. The mass of each tooth (zeroth moment of the particle distribution, $M_0$) is seen to be one-to-one with $\phi_1$; higher-order moments  ($M_K,~k=1,\ldots,5$) appear slaved to $\phi_1$. The blue dashed line shows the slaving function for uniform distribution. }
 \label{fig_variableID}
\end{figure*}

When the micro-scale description is in the form of particles (or agents), the macro-scale variable is expected to represent some local statistics of the particles. In other words, we have to identify the statistical variables that lead to a `closed' set of governing equations for the system. 
To capture such variables, we take snapshots of the gap-tooth simulation and consider the particle distribution in each tooth as a single data point; this results in a cloud of data points in a space of distributions (see \cref{fig_variableID}(a)). We hypothesize that this cloud from our simulations lies close to a low-dimensional manifold. We use Diffusion Maps \cite{coifman2006diffusion} to test our hypothesis and discover the dimensionality, as well as a parameterization of this manifold. This parameterization (i.e., the Diffusion-Map coordinate(s)), will be our candidate macro-scale variable(s). In previous works, we have successfully identified and used such coordinates  \cite{erban2007variable}, namiing the procedure ``variable-free computation". To learn the structure of the manifold, we need to define a notion of distance between our data points. Given that each data point is a particle distribution, this amounts to selecting a notion of distance between such distributions that can be robustly estimated from data.

\subsection{Unnormalized Optimal Transport Distance}
 An emerging paradigm in machine learning/applied probability that focuses on computing the distance between distributions is  \emph{optimal transport} theory \cite{villani2008optimal,peyre2019computational}: the distance between two distributions is defined as the cost of optimally `moving' one distribution to another. 
 This type of distance in the context of manifold learning has been recently used for characterizing molecular conformations in proteins \cite{zelesko2020earthmover}. 
The bulk of optimal transport theory is concerned with probability (i.e. unit total mass) distributions; here we are interested in distances between particle distributions with varying total masses. Hence we utilize  more recent formulations of this theory which discuss transport distances between `unbalanced' or `unnormalized' distributions \cite{chizat2018scaling,gangbo2019unnormalized}.

We use the \emph{unnormalized optimal transport} formulation in \cite{gangbo2019unnormalized} to compute distances between particle distributions in each tooth. Let $\mu_1$ and $\mu_2$  be particle densities within two equally sized teeth; consider them supported on the same one-dimensional interval $I\subset \mathbb{R}$. 
Following the setup in \cite{gangbo2019unnormalized}, we assume $\mu_1$ and $\mu_2$ are non-negative integrable functions. 
The transport formulation is based on finding a \emph{time-dependent flow with a source term}  that evolves the distribution from $\mu_1$  to $\mu_2$ over the time interval $[0,1]$. The $L^p$ \emph{unnormalized Wasserstein distance} between these two distributions is defined as
\begin{align}\label{eq_UWP}
d_{UW_p}(\mu_1,\mu_2)^p&=\inf_{\mu,v,f}\int_{0}^{1}\int_I \|v(t,x)\|^p \mu(t,x)dxdt  \notag
\\ &+\beta\int_0^1|f(t)|^pdt |I| 
\end{align}
subject to the constraints
\begin{align}\label{eq_UWcons} 
\p_t\mu(t,x)+\nabla (\mu(t,x)v(t,x))=f(t), \\ \mu(0,x)=\mu_1(x) ~~\text{and} ~~  \mu(1,x)=\mu_2(x).
\end{align}
The infimum in \cref{eq_UWP} is taken over all choices of  density function $\mu$  defined on $[0,1]\times I$, the vector field $v$ defined on $[0,1]\times I$ and with zero flux on the boundaries of $I$; the source function $f$ is defined on $[0,1]$. Here $|I|$ is the length of the tooth.
Interestingly, if $I=[0,1]$ and $p=1$, an explicit expression exists for the unnormalized Wasserstein distance:
\begin{align}\label{eq_UW1}
d_{W}\myeq d_{UW_1}(\mu_1,\mu_2)=& \int_I \bigg| \int_{0}^x \big(\mu_1(z)-\mu_2(z)\big) dz \notag
\\  & - x\int_I \big(\mu_1(z)-\mu_2(z)\big)dz \bigg| dx \notag
 \\ &+ \beta \left| \int_I \big(\mu_1(z)-\mu_2(z)\big)dz\right|.
\end{align}

\Cref{fig_variableID}(b, top left) shows the unnormalized Wasserstein distance computed across pairs of particle distributions within teeth of gap-tooth simulation snapshots.  These pairwise distances capture the regularity in the macro-scale behavior of the distributions: teeth close in space-time have distributions that close to each other in the transport distance. This is reflected in the low values of entries close to the diagonal of the distance matrix, and its anti-diagonal for a periodic domain.
In our one-dimensional example, this formulation of transport is intuitive and easy to verify. However, in \cref{app_distances}, we present pairwise distance matrices computed using the unbalanced optimal transport formulation in \cite{chizat2018scaling} as well as a non-transport based, moments formulation. Those formulations yield the same \emph{qualitative} pairwise distance matrix structure.

\subsection{Diffusion Maps}
Equipped with a notion of distance between particle distributions, we use Diffusion Maps  \cite{coifman2006diffusion,nadler2006diffusion} to discover the dimensionality, and a  parameterization of the manifold  formed by the ensemble of particle distributions. The main idea is to approximate the diffusion operator  on the manifold (the Laplace-Beltrami operator) and find its eigenfunctions. One can use a subset of the leading independent eigenfunctions, called \emph{Diffusion-Map coordinates}, as a geometric parameterization of the manifold  \cite{erban2007variable,singer2009detecting}.

Let $\mu_j,~j=1,2,\ldots,m$ be the set of particle distributions within the teeth from a representative snapshot of gap-tooth simulations. To compute Diffusion-Map coordinates we first construct the diffusion kernel matrix ${W}$,
\begin{align} \label{eq_diffkernel}
W_{ij}=\exp\left(-\frac{d_W(\mu_i,\mu_j)^2 }{\epsilon}\right)
\end{align}
where $d_W$ is the transport distance defined in \cref{eq_UW1} and $\epsilon$ is the diffusion kernel width. The value of $\epsilon$ is chosen in accord with the target length scales of the manifold. Here we use a common heuristic approach, choosing $\epsilon$ to be the median of the distance values for our data points. 
In the next step, we mitigate the effect of non-uniform sampling on the manifold by an appropriate normalization of the kernel matrix,
\begin{align}
\overline{{W}}= {D}^{-1}{W}{D}^{-1}
\end{align}
where ${D}$ is a diagonal matrix with $i$-th entry the sum of the $i$-th row in ${W}$. This matrix is shown in \cref{fig_variableID}(b, bottom left). The kernel matrix has to be normalized again to be row stochastic, i.e., 
\begin{align}
\hat{{W}}= \overline{{D}}^{-1}\overline{{W}}.
\end{align}
where $\overline{D}$ is a diagonal matrix with row-sums of $\overline{{W}}$ on its diagonal.
The diffusion operator on the manifold is approximated by 
\begin{align}
A={I}-\hat{{W}}.
\end{align}
with $I$ being the identity matrix.
 The Diffusion-Maps eigenfunctions are given as eigenvectors of this matrix,
\begin{align}
{A}\phi_k=\lambda_k\phi_k, \quad k=0,1,\ldots,m.
\end{align}
A minimal subset of these eigenfunctions which are mutually independent form the parameterization for the manifold. This can be accomplished by inspecting the statistical independence of computed eigenfunctions \cite[e.g.][]{dsilva2018parsimonious}.

Application of the Diffusion-Maps algorithm, with the unnormalized optimal transport distance in the diffusion kernel, to snapshots of Burgers gap-tooth simulation shows that there exists only one independent Diffusion-Maps coordinate, denoted by $\phi_1$, for the particle distributions in all the teeth of these  snapshots (see \cref{app_distances}).  This confirms our hypothesis that local particle distributions lie close to a low-dimensional (here one-dimensional) manifold in the space of distributions; moreover, the macro-scale spatial behavior of the particle distributions is well represented by just one macro-scale variable.  
 \Cref{fig_variableID}(b, right) shows that, as expected, the data-driven variable $\phi_1$ is one-to-one with the zeroth moment (i.e. mass) of particle distributions, and hence one-to-one with the local particle density field value.  
 Notice in \cref{fig_variableID}(b) that higher-order particle distribution moments are also one-to-one with the leading diffusion map coordinate (and thus slaved to the local density field). As the figure shows,  this slaving is consistent with locally  uniform distributions.

\section{Learning PDEs via neural nets \label{sec_learning}}
Given a right choice for the macro-scale variable, either motivated by physical considerations or discovered from data, we use machine learning techniques to learn the right-hand-side of the PDE governing the evolution of that variable. We denote this variable here by $v$: either the particle density field, $\rho$, or the data-driven variable $\phi_1$ discovered in the last section.

We consider two general architectures for learning the macro-scale PDE through feedforward neural networks. In the first architecture, we look for \emph{the PDE law} in the functional form
\begin{equation}\label{eq_PDE1}
v_t=F\big(v,v_x,v_{xx}, \ldots).
\end{equation}
The number of spatial derivatives appearing in the PDE right-hand-side is typically unknown beforehand. In previous work we presented data-driven methods, based on manifold learning and Gaussian process regression, for testing whether different subsets of spatial derivatives are sufficient to learn the time evolution \cite{lee2020coarse}.
Given such a sufficiently rich subset, learning in this architecture is equivalent to regressing the function $F$ from data on these derivative terms.

The second architecture represents the unknown PDE in its spatially discretized form, i.e.,
 \begin{equation}\label{eq_PDE2}
\frac{dv^j}{dt}=G\big(v^j,v^{j-1},v^{j+1}, \ldots).
\end{equation}
where $v^j$ is the value of $v$ at $x^j$, a point on a local spatial grid ordered by index $j$. The stencil size of the discretization (i.e. the number of neighboring grid points required to approximate the spatial derivatives required to learn the time derivative) depends on the number of spatial derivatives involved (and the regularity of the PDE solution) and can be tested as described in \cite{lee2020coarse}. This general type of architecture has been demonstrated to suffice for capturing the nonlinearity of the Burgers as well as a few other PDE laws in \cite{bar2019learning}.

\begin{figure}
\centering
\subfloat[][]{\includegraphics[scale=1]{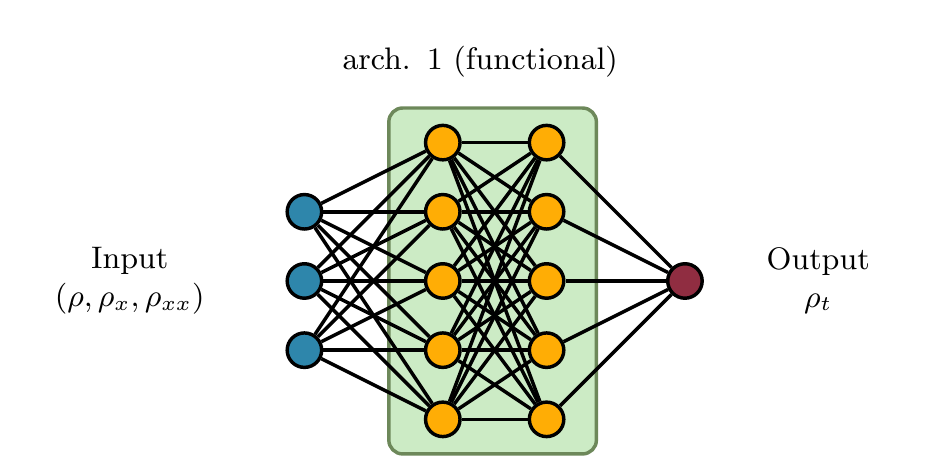}} 

\subfloat[][]{\includegraphics[scale=1]{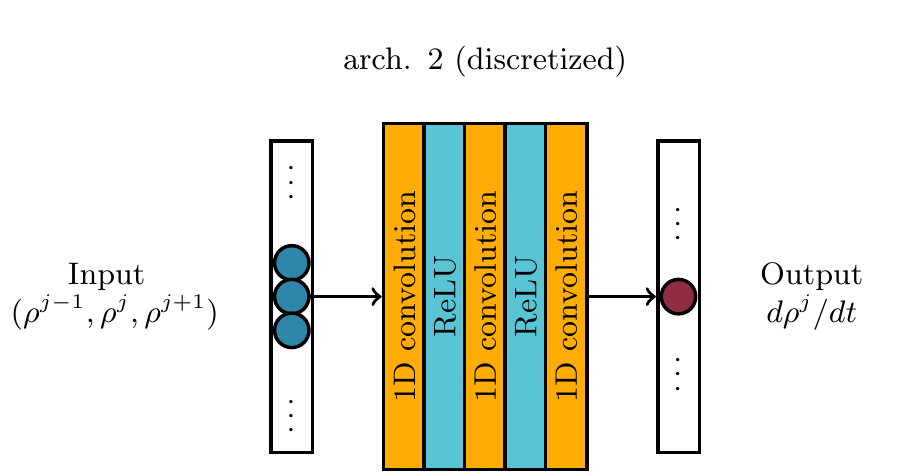}}

\caption{\textbf{Neural network architectures for learning the macro-scale PDE.} a) The neural net architecture for learning the PDE in the form of  \eqref{eq_PDE1}: There are two fully connected layers with 48 nodes; the first  layer has ReLU nonlinearity. b) The neural net architecture for learning the PDE in the form of  \eqref{eq_PDE2}: There are three convolutional layers with 48 filters and ReLU activations in between. This network learns the \emph{discretized} PDE from values of the solution on local stencils. }
\label{fig_nnarch}
\end{figure}

 We use a feedforward neural network architecture to regress $F$, and a convolutional neural network to regress $G$. For the learning of the PDE with density as dependent variable, the first network consists of two fully connected layers, each with 48 nodes, and a Rectifying Linear Unit (ReLU) layer in between. 
The second network consists of three convolutional layers, each having 48 filters, with ReLU activations in between.  These two networks are shown in \cref{fig_nnarch}. For the data-driven dependent variable, the first architecture has three fully connected layers with 64 nodes, while the second architecture has four convolutional layers with 64 filters.

In previous work we successfully utilized such architectures to learn coarse-grained PDEs from data on detailed simulations of multiscale diffusion in materials with micro-scale heterogeneity \cite{arbabi2020linking}. The specific architecture (i.e. number of layers/nodes) here is determined by a local grid search minimization of error starting from the architectures in \cite{arbabi2020linking}; architectures with up to a few more layers were also found adequate.
We implement all networks in TensorFlow 2.0  \cite{tensorflow2015-whitepaper} and  train them using the ADAM optimizer  \cite{kingma2014adam}. Each network is trained by minimizing the Mean Squared Error (MSE) of the  time-derivative predictions.

\section{Results \label{sec_results}}
We demonstrate the application of our learning framework in the particle model introduced in \cref{sec_setup}. To generate data, we use the gap-tooth scheme with parameter values $Z=5\times 10^5$, $\alpha=0.1$, $N=128$ and $\nu=0.05$. We simulate 12 two-second-long trajectories with initial conditions of the form
\begin{equation}\label{eq_IC}
\rho_0(x)=\sum_{k=1}^{20}A_k\sin(l_k x + \phi_k)
\end{equation}
where $A_k$, $l_k$ and $\phi_k$ are drawn randomly from uniform distributions on $[-.5,.5]$, $\{1,2,\ldots,7\}$ and $[0,2\pi)$, respectively. We record snapshots of $\rho$ at intervals of length $10^{-3}$ and compute associated snapshots of $\p_t \rho$ using temporal finite differences. This results in 12000 pairs of $(\rho,\p_t\rho)$ snapshots. Note that, using numerical integrators as templates for the neural network architecture, the right-hand-side of the PDE could be learned from discrete data without time derivative estimation \cite{gonzalez1998identification}.
We use data from 8 trajectories to train the network, and two more trajectories as a validation dataset to optimize the number of layers/nodes in each architecture. The remaining two trajectories constitute our test data.

We learned the data-driven variable $\phi_1$ from a few snapshots of gap-tooth simulation; we can transform other data snapshots using out-of-sample extension methods from manifold learning (Nystr\"om extension and Geometric Harmonics \cite{coifman2006geometric}). Here, we simply learn the functional relationship between $\rho$ and $\phi_1$ and regress the values of $\phi_1$ for new snapshots. By applying this to the above trajectories, we obtain an equal number of $(\phi_1,\p_t\phi_1)$ snapshots to train, validate and test the neural networks for learning the PDE of this variable.

\begin{figure*}
\centering
\subfloat[][]{\includegraphics[scale=1]{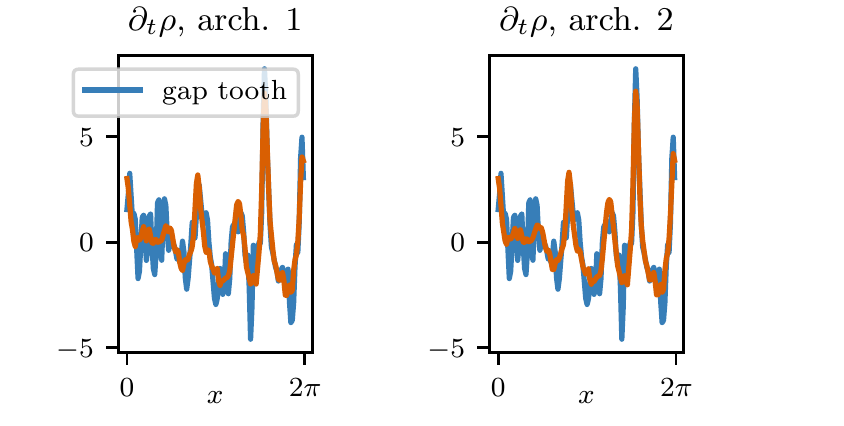}} 
\subfloat[][]{\includegraphics[scale=1]{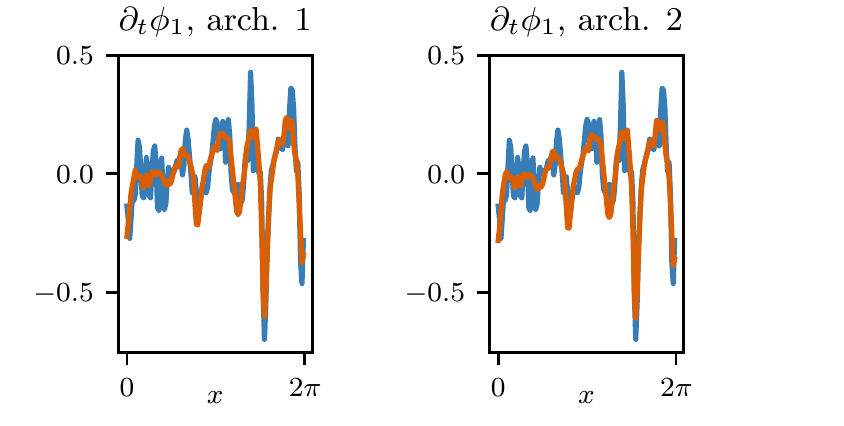}} 

\subfloat[][]{\includegraphics[scale=1]{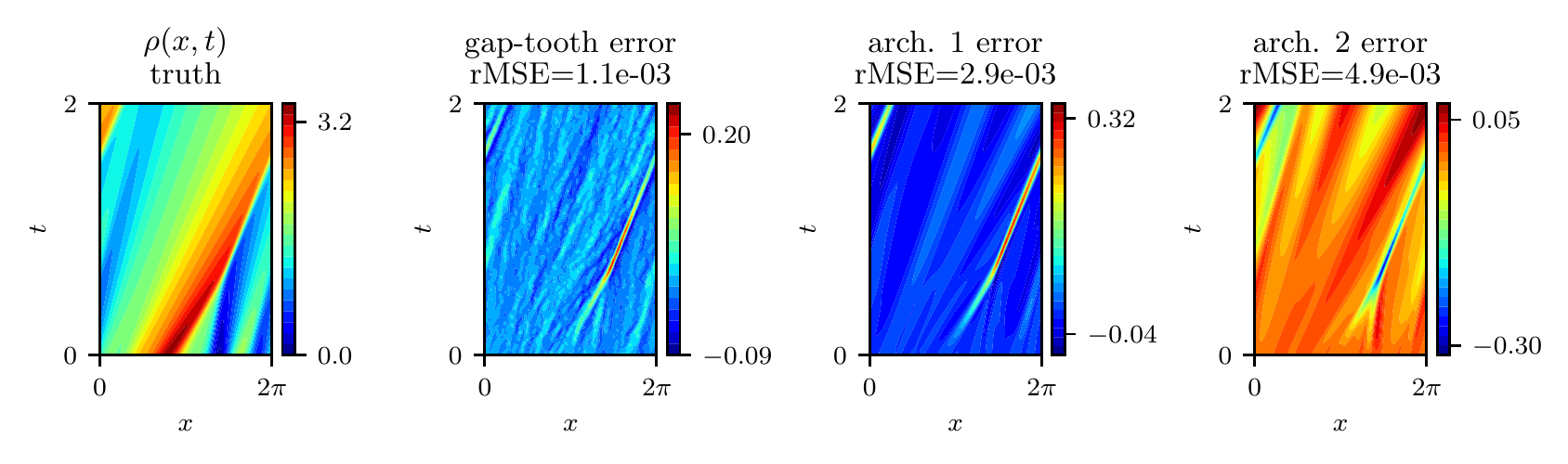}}

\subfloat[][]{\includegraphics[scale=1]{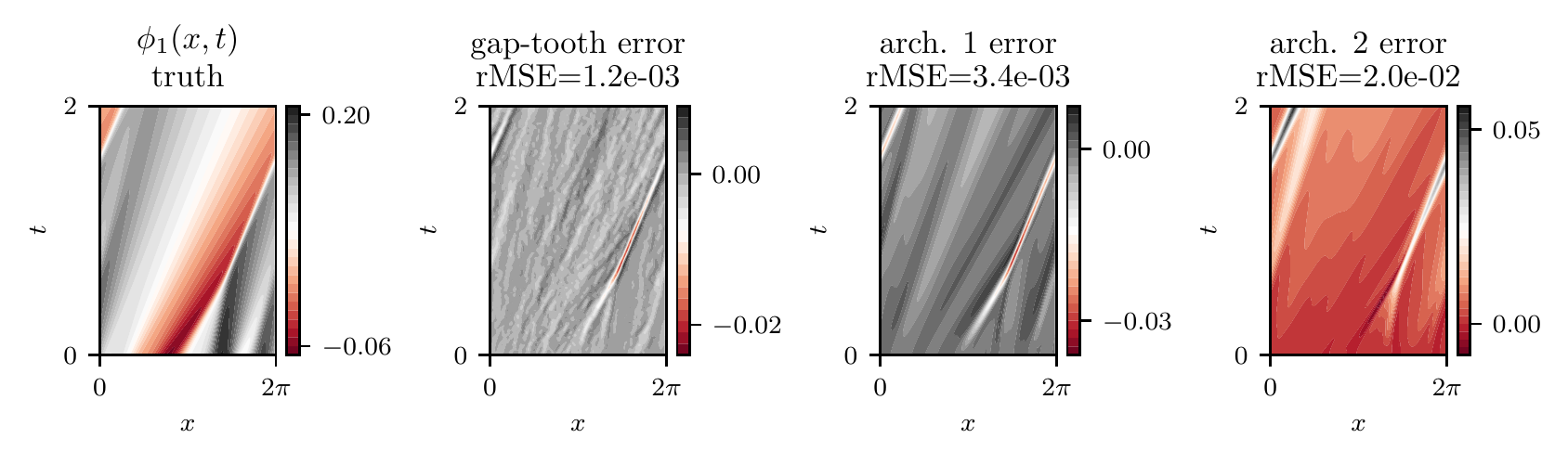}} 
 
\caption{\textbf{Performance of trained models.} a,b) The neural nets (trained with noisy data) predict a smoothed version of the $\p_t \rho$.  c,d) Using the trained neural nets inside a time stepper leads to trajectory predictions with accuracy close to the gap-tooth scheme, both for the physical variable $\rho$ and the data-driven variable $\phi_1$.}
\label{fig_result}
\end{figure*}

The scalar macro-scale dependent variable (density or $\phi_1$) represents a statistical feature of the particles; fine scale simulations reported as density (or $\phi_1$) values, would carry stochastic fluctuations. These fluctuations are amplified when estimating the spatial and temporal derivatives, resulting in noisy data for training and testing: \cref{fig_result}(a,b) shows the time-derivative of density and the data-driven variable for a test snapshot and predictions made by the neural nets. 
One approach to mitigating this noise is to address stochasticity by pre-smoothing the data. 
The neural nets tend to predict smooth snapshots even when trained with noisy outputs. To rationalize this, recall the properties of optimal predictors \cite[e.g.][]{friedman2001elements}: if the training data has multiple identical inputs with different outputs due to noise, the model, choosing MSE as the loss function, will predict the average of those outputs for the same input, in effect removing the (unbiased) noise. The caveat for neural networks is that their large capacity allows them to overfit the noise for nearby inputs; we believe that our networks are sufficiently regularized (i.e., shallow) so as to be robust to noise in the training data in line with the above explanation.

We use the trained networks to integrate test trajectories of the learned PDEs for density and for the data-driven variable. \Cref{fig_result}(c,d) shows an example of such a trajectory, with an unseen initial condition, compared to the gap-tooth simulation, and a high-order finite-volume solution \cite{shu2009high}  of the Burgers PDE taken as ground truth. The results show that neural networks achieve an accurate and stable estimation of the PDEs, leading to trajectory prediction accuracy close to the particle gap-tooth scheme itself.  
The $\phi_1$ predictions, similar to $\rho$, can be lifted back to positions of particles. Beyond traditional ``lifting" in  equation-free numerics, \cite{kevrekidis2003equation} one can use modern ML generative methods, including GANs \cite{goodfellow2014generative}, for this purpose.

\section{Conclusion}
In this work, we discussed two remedies for difficulties of learning effective coarse-grained PDEs from microscopic data: use of multi-scale numerics to reduce the burden of sufficient data generation for training; and manifold learning on fine-scale simulation snapshot data to extract the dependent variable(s) of the PDE.
In particular, we used the notion of unnormalized Wasserstein distance in discovering the macro-scale data-driven (dependent) variable for our coarse-grained PDE. A natural next step is to extend the variable identification framework to systems with multiple particle species, and the right macro-scale variable(s) should encode information about the particle distributions but also their interactions. This task  requires a new notion of distribution distance, which is additionally informed by physics at the microscopic level. Finally, we note that one could even infer the {\em independent} variables (the appropriate parameterizations of space and even time) for coarse-grained PDEs in systems where there is no \emph{a priori} notion of physical space or time for modeling \cite{kemeth2018emergent,arbabi2020coarse,kemeth2020learning}.

\begin{acknowledgments}
This work was partially supported by the US A.R.O. through a MURI and by DARPA. The authors are not aware of any conflicts of interest in publishing this work.
\end{acknowledgments}

\section*{source code and data}
Source codes for implementation of our framework and reproducing the results of this paper are available at  \url{https://github.com/arbabiha/Particles2PDEs}.

\clearpage

\appendix
\section{Gap-tooth boundary conditions for particle simulations}\label{app_gaptooth}
For completeness, we briefly recall the implementation of boundary conditions in our gap-tooth scheme following \cite{gear2003gap}.
Consider the gap-tooth setup and grid for the particle Burgers model described in \cref{sec_gaptooth}. 
Recall that we lift the initial density profile $\rho_0$ to a particle state within each tooth and  simulate the microscopic dynamics in \eqref{eq_Burgersmicro} for all the particles over a short time period. The particles that exit each tooth are redirected to the neighboring teeth using the flux redistribution laws. These laws are obtained by interpolating  the influx at one tooth from the outfluxes at  the neighboring teeth. 
More precisely, let $I_{r,i}$ be the right-going influx, and $O_{r,i}$ be the right-going outflux at the $i$-th tooth. Note that $I_{r,i}$ crosses the left boundary of the tooth while $O_{r,i}$ crosses the right boundary as shown in \cref{fig_gaptooth}(b). Using a quadratic interpolation, we can compute the influx as 
\begin{equation}\label{eq_BC2}
I_{r,i}=\frac{\alpha(1+\alpha)}{2}O_{r,i-1}+(1-\alpha^2)O_{r,i}-\frac{\alpha (1-\alpha)}{2}O_{r,i+1}.
\end{equation}
Although this boundary condition for each tooth is simple, to interpret and implement it for particle dynamics, it is better to examine how an outflux gets redistributed to influxes. Using the same quadratic formula we see that the outflux  $O_{r,i}$ is split into three parts: $(1-\alpha)^2$ fraction of particles go to $I_{r,i}$ (i.e. redirected to the same tooth), $\alpha(1+\alpha)/2$ fraction go to $I_{r,i+1}$ (i.e. jump into the downstream tooth) and $-\alpha(1-\alpha)$ fraction are redirected to  $I_{r,i-1}$ (i.e. jump to the upstream tooth). 
 Interestingly, the last fraction is negative! In the gap-tooth scheme, this negative fraction is realized as \emph{anti-particles} and \emph{annihilation}: we count the outgoing particles, and generate anti-particles that equal $\alpha(1-\alpha)$ fraction of those outgoing particles. We place those anti-particles inside the upstream tooth and cancel them out with the closest particles. This implementation preserves the total number of particles in the system through time. In addition, if a particle jumps the distance $\delta$ outside a tooth, it (or its anti-particle) will be inserted at distance $\delta$ into the receiving tooth. The left-going fluxes are treated similar to the right-going fluxes with the notion of upstream and downstream teeth are appropriately adjusted.
The redistribution of the outfluxes is repeated until all particles are positioned properly inside a tooth and all anti-particles are annihilated. 

 \begin{figure}
\centering

\includegraphics[scale=1]{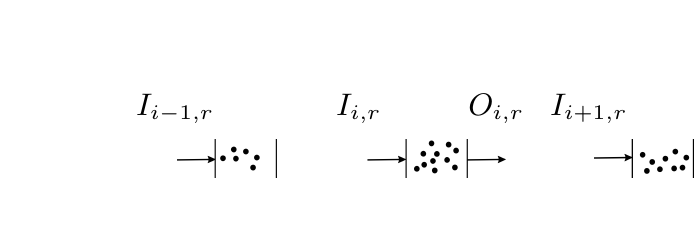}
\caption{\textbf{Tooth boundary conditions in gap-tooth scheme} The flux redistribution for particles exiting a tooth: the outflux of particles from a tooth is redirected to itself and the neighboring tooth.}
\label{fig_gaptooth_BC}
\end{figure}

\section{Notions of distance between distributions}\label{app_distances}
We discussed the notion of the unnormalized optimal transport distance \cite{gangbo2019unnormalized} as the metric used in the Diffusion Maps kernel for discovery of appropriate macro-scale variable(s)  from data. Here we review two other notions for distance: moments distance and unbalanced optimal transport \cite{chizat2018scaling}. 
\subsection{Moments distance}
Let $\mu$ denote a particle distribution density on an interval $I\subset \mathbb{R}$ .  The $k$-th moment of this distribution is defined as
\begin{align}
M_k(\mu)=\int_{I}x^k\mu(x)dx.
\end{align}
We can also approximate the moments directly from the position of particles, that is,
\begin{align}
M_k(\mu)\approx\frac{1}{n}\sum_{p=1}^{n} x_p^k
\end{align}
where $x_p$ is the position of the $p$-th particle.
It turns out that if the interval $I$ is compact, knowing the sequence of moments $\left\{M_k(\mu)\right\}_{k=0}^\infty$ uniquely distinguishes $\mu$ from other distributions \cite{schmudgen2017moment}. In other words we can use the moments to represent each distribution without any loss of information. 

Here, we use a finite truncation of a moment representation to compute the distance between particle distributions on the same-sized intervals.  
First, we define a $K$-term truncation of the moments for each distribution,
\begin{align}
\tilde{M}(\mu) = \left[M_0(\mu),~M_1(\mu),\ldots,~M_K(\mu) \right]^\top
\end{align}
We define \emph{the moment distance} of two distributions, $\mu_i$ and $\mu_j$, as the Euclidean distance between their truncated moment vectors:
\begin{align} \label{eq_momdist}
d_M(\mu_i,\mu_j) \myeq \left\| \tilde{M}(\mu_i) -\tilde{M}(\mu_j)\right\|
\end{align}

\subsection{Unbalanced optimal transport}
Let $\mu_1$ and $\mu_2$  be two particle distribution densities on the interval $I$. Also let $c:I\times I\rar \mathbb{R}^+$ be a positive and continuous cost function for transport in $I$, that is, the cost of moving one unit of mass from $x_1$ to $x_2$ is $c(x_1,x_2)$. We are interested in finding the plan for moving $\mu_1$ to $\mu_2$ that minimizes the overall cost of this moving using the cost function $c$. More importantly, we want to compute the minimal cost obtained by this plan as a distance between the two distributions. Here we review and utilize the formulation in \cite{chizat2018scaling}.

To make the transport problem more precise, let $\M^+(I,I)$ be the space of (positive) joint distributions on the product of $I$ and itself.  For any $\gamma \in \M^+(I,I)$, we define $P^1_\#(\gamma)$ to be its first marginal and $P^2_\#(\gamma)$ its second marginal.  The classical optimal transport problem is then formulated as \cite{chizat2018scaling}:
\begin{align}
\min_{\gamma \in \M^+(I,I)} \mathcal{I} (\gamma) \myeq\int_{I\times I} c(x,y) \gamma(x,y)dxdy\qquad 
\\  \text{s.t.} \quad P^1_\#(\gamma)=\mu_1 ~\text{and} ~P^2_\#(\gamma)=\mu_2.
\end{align}
The term $\gamma(x_1,x_2)$ can be interpreted as the amount of density moved between $x_1$ and $x_2$, and therefore the integral term is in fact the total cost of  transport. On the other hand, the constraints require that marginals of $\gamma$ be exactly $\mu_1$ and $\mu_2$. In other words, the total amount of density shipped from $x_1$  should be $\mu_1(x_1)$ and the total amount received at $x_2$ should be $\mu_2(x_2)$. If $\mu_1$ and $\mu_2$ have different masses, i.e., the transport is unbalanced, the optimization problem becomes infeasible because marginals of a joint distribution have the same mass. The formulation of unbalanced optimal transport in \cite{chizat2018scaling} relies on replacing the hard constraints on marginals with soft constraints. 
Recall that Kullback--Leibler (KL) divergence defined as
\begin{align}
\D_{KL}(\mu_1||\mu_2) \myeq \int_{X} \mu_1(x)\log\bigg(\frac{\mu_1(x)}{\mu_2(x)}\bigg)dx.
\end{align}
The unbalanced optimal transport in \cite{chizat2018scaling} is formulated as
\begin{align}\label{eq_UOT}
\min_{\gamma \in \M^+(I,I)} \J (\gamma) &\myeq\int_{I\times I} c(x,y) \gamma(x,y)dxdy \notag
\\ &+\alpha \D_{KL}\left(P^1_\#(\gamma)||\mu_1\right) + \alpha \D_{KL}\left(P^2_\#(\gamma)||\mu_2\right).
\end{align}
The parameter $\alpha$ determines how much the violation of marginals is penalized in realizing the optimal transport plan. The minimal value of the objective function above is the optimal transport distance between the two unbalanced distributions. To distinguish this distance from other distances used in this work we use the following notation:
\begin{align}\label{eq_WFR}
d_{U}=  \min_{\gamma \in \M^+(I,I)} \J (\gamma).
\end{align}
 
The work in  \cite{chizat2018scaling} uses entropic regularization, to derive an iterative algorithm for finding the optimal plan (and optimal transport distance)  which has a closed analytical form for each iteration. This makes the algorithm well-suited for large datasets, and here we use it to compute the pairwise distance between distributions from the gap-tooth simulations.
We refer the reader to \cite{chizat2018scaling} for an extensive analysis of the formulation and the numerical algorithm.

\subsection{Application to a gap-tooth snapshot}
\begin{figure*}
\centering
\includegraphics[scale=1]{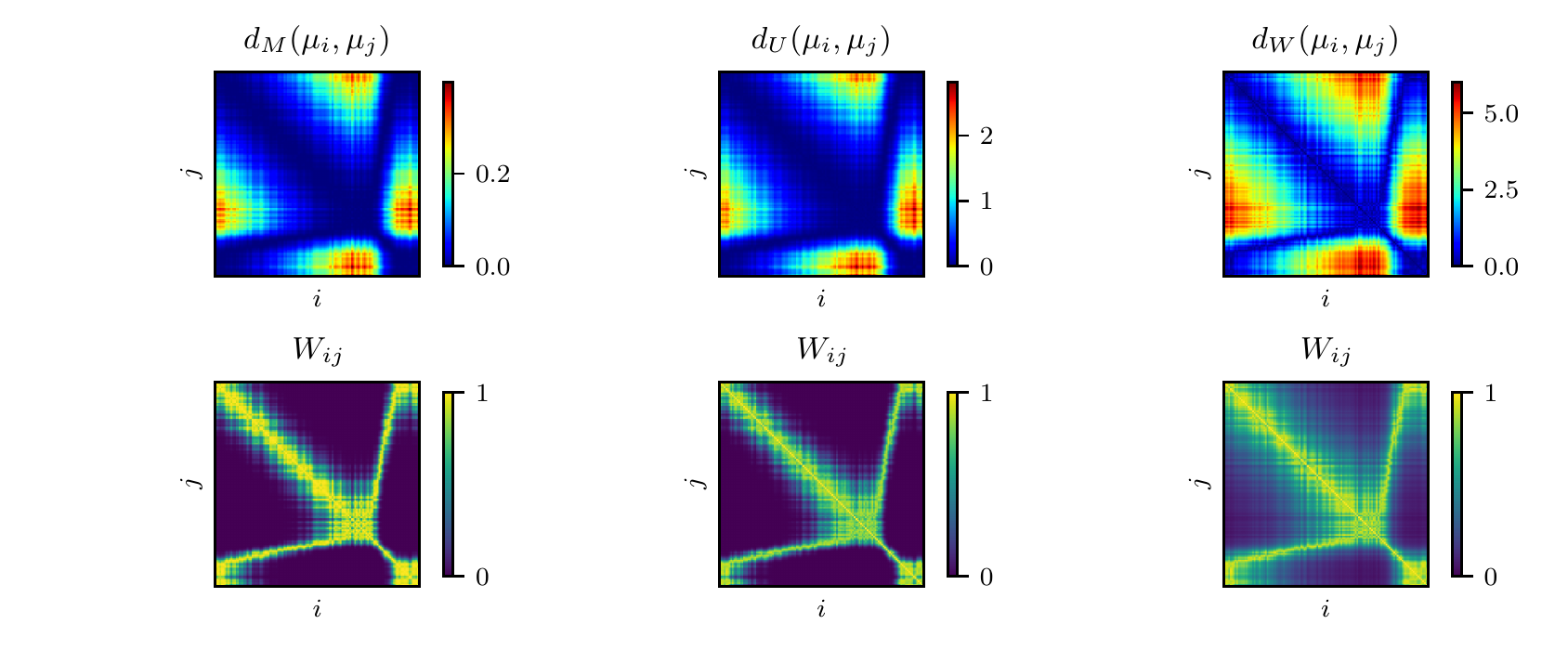}

\caption{ \textbf{Pairwise distance of particle distributions in a snapshot of gap-tooth simulation:} Top left: the moments distance \cref{eq_momdist}, top middle: unbalanced optimal transport distance \cref{eq_UOT} \cite{chizat2018scaling}, top right: unnormalized optimal transport \eqref{eq_UWP} \cite{gangbo2019unnormalized}. The bottom row shows the corresponding diffusion kernel \eqref{eq_diffkernel}. }
 \label{fig_distances}
\end{figure*}

\Cref{fig_distances} show the pairwise distances for particle distributions inside the teeth in a snapshot of the Burgers gap-tooth simulation. The snapshot was shown in \cref{fig_variableID}(a). \textbf{ The different formulations of transport distance and moments distance lead to the same qualitative structure in the distance matrix and diffusion kernel}.

\begin{figure*}
\centering
\includegraphics[scale=.9]{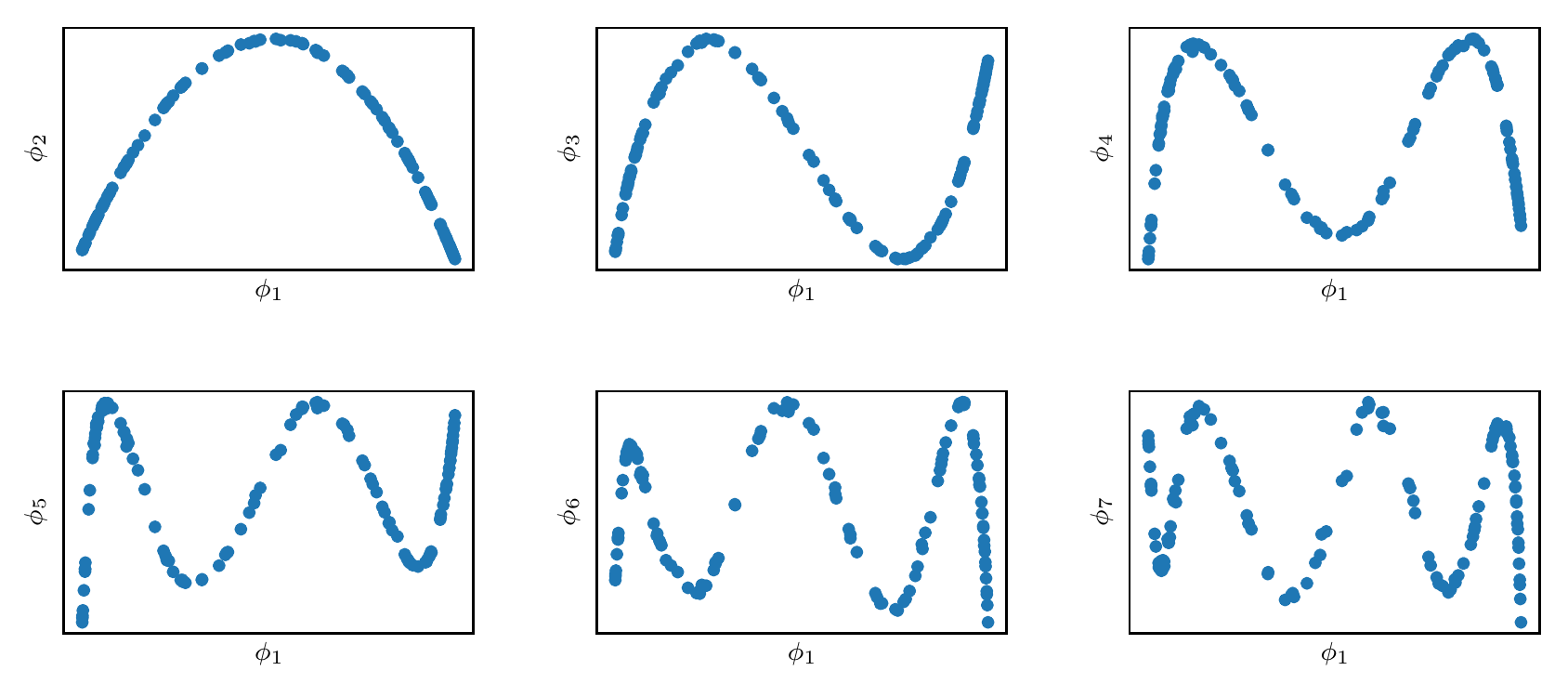}

\caption{ \textbf{Inter-dependence of the Diffusion-Maps coordinates:} The scatter plots of second to seventh dominant coordinate vs. the first one, computed using the pairwise distances of unnormalized optimal transport. The coordinates are dependent on $\phi_1$  showing that data manifold in the space of distributions is (approximately) one dimensional. }
 \label{fig_dmaps}
\end{figure*}

\Cref{fig_dmaps} shows the scatter plots of the second to seventh dominant Diffusion-Map coordinates vs the first dominant Diffusion-Map coordinate, $\phi_1$, which was chosen as the data-driven variable for the coarse-grained PDE. The figure shows that those coordinates are highly dependent on $\phi_1$, therefore the data points lie close to a one-dimensional manifold in the space of distributions, and $\phi_1$ gives a parameterization of that manifold.

\bibliography{main.bbl}

\end{document}